\documentclass{IOS-Book-Article}

\usepackage{mathptmx}
\usepackage{graphicx}
\usepackage{subcaption}
\usepackage[export]{adjustbox}
\usepackage{wrapfig}
\usepackage{amsmath,amssymb,amsfonts}
\usepackage{mathtools}
\usepackage{comment}
\usepackage{multicol}
\def\hb{\hbox to 10.7 cm{}}

\begin{document}

\pagestyle{headings}
\def\thepage{}

\begin{frontmatter}              % The preamble begins here.

%\pretitle{Pretitle}
\title{Evaluating the Usefulness of Unsupervised monitoring in Cultural Heritage Monuments}

\markboth{}{May 2021\hb}
%\subtitle{Subtitle}

\author[A]{Charalampos Zafeiropoulos}
\author[A]{Ioannis N. Tzortzis}
\author[A]{Ioannis Rallis}
\author[A]{Eftychios Protopapadakis}
\author[A]{Nikolaos Doulamis}
\author[A]{Anastasios Doulamis}

\address[A]{National Technological University of Athens, \{mpampiszafeiropoulos, itzortzis, eftprot \}@mail.ntua.gr, irallis@central.ntua.gr, \{adoulam, ndoulam\}@cs.ntua.gr }

%\runningauthor{C. Zafeiropoulos et al.}

%\author[]{XXXXXXXXXXXXXXXXXXXXXXXX}
%\author[]{XXXXXXXXXXXXXXXXXXXXXXXX}
%\author[]{XXXXXXXXXXXXXXXXXXXXXXXX}
%\author[]{XXXXXXXXXXXXXXXXXXXXXXXX}
%\author[]{XXXXXXXXXXXXXXXXXXXXXXXX}
%\author[]{XXXXXXXXXXXXXXXXXXXXXXXX}

%\runningauthor{XXXXXXXXXXXXXXXXXXXXXXXX et al.}

\begin{abstract}
In this paper, we scrutinize the effectiveness of various clustering techniques, investigating their applicability in Cultural Heritage monitoring applications. In the context of this paper, we detect the level of decomposition and corrosion on the walls of Saint Nicholas fort in Rhodes utilizing hyperspectral images.
%Cultural heritage assets denote the "modern" history of each country. Therefore, their protection, monitoring and maintenance constitute a major challenge for global organizations such as UNESCO as well as for the scientific world. In the context of this paper, we detect the level of decomposition and corrosion on the walls of Saint Nicholas fort in Rhodes.
A total of 6 different clustering approaches have been evaluated over a set of 14 different orthorectified hyperspectral images.
%Then, we proposed an unsupervised machine learning technique called clustering on the hyperspectral images to detect any decomposition and/or corrosion.
Experimental setup in this study involves K-means, Spectral, Meanshift, DBSCAN, Birch and Optics algorithms.
For each of these techniques we evaluate its performance by the use of performance metrics such as Calinski-Harabasz, Davies-Bouldin indexes and Silhouette value. In this approach, we evaluate the outcomes of the clustering methods by comparing them with a set of annotated images which denotes the ground truth regarding the decomposition and/or corrosion area of the original images. The results depict that a few clustering techniques applied on the given dataset succeeded decent accuracy, precision, recall and f1 scores. Eventually, it was observed that the deterioration was detected quite accurately.
\end{abstract}

\end{frontmatter}
\markboth{May 2021\hb}{May 2021\hb}

\section{Introduction}
\label{Introduction}

\indent 
Cultural Heritage assets (monuments, artefacts and sites) suffer from on-going deterioration through natural disasters, climate change and human negligence or interventions. Monuments are defined as structures created by a person or event and they symbolize a historic period of the corresponding place due to its artistic \cite{rallis2017extraction}, historical, political, technical or architectural importance \cite{moropoulou2013non}. UNESCO considers as a first priority the preservation and valorisation of the tangible/intangible Cultural Heritage and applies innovative techniques for the capturing, digitizing, documenting and preserving prestigious monuments \cite{marrie2008unesco}, \cite{adamopoulos2017multi}. 

Early detection of decay and deterioration is essential to preserve monuments. Material degradation leads to the failure of the the buildings components. Non-destructive techniques utilized for detection of monument decay. Most of these techniques come from the scientific fields of computer vision, whose great goal is the extraction of information regarding regions of interest (ROIs) from images or sequences of images \cite{forsyth2012computer}. 

The state-of-the-art for 3D/4D documentation and modelling of complex sites utilizes multiple sensors and  technologies (e.g., LIDAR, photogrammetry, in-situ surveying, hyperspectral sensors) in order to define the preservation status of the monument\cite{maltezos2018building}. The efficient use of these tools can give significantly better material detection and object recognition, and thus to identify even the smallest differences in spectral signatures of various objects. This continuous development of new sensors, data capturing methodologies, computer vision algorithms,  multi-resolution 3D/4D representations and the improvements of existing ones are contributing significantly to the growth of the interdisciplinary cultural heritage domain. 

Hyperspectral images are more suitable than RGB ones since they provide a large amount of information (high-spatial and high spectral resolution), allowing identifying the screened materials based on their chemical composition rather than only their size, shape, and visible colour \cite{makantasis2015deep}. Moreover, the recent advancement of sensors technologies has led to the development of hyperspectral imaging sensors with higher spectral and spatial resolution on-board various satellite, aerial, UAV and ground acquisition platforms.

%Most hyperspectral data classification methods consist of two steps. First step defines the features created form raw data, and the second step includes the features which are used to create classifiers such as Support Vector Machines (SVM) and Neural Networks (NN). In order to face high dimensional data, we use statistical learning methods. However, due to high variety of materials, is difficult to know which features are important for the classification. In general, deep learning models seem to succeed more efficient classification on big datasets and large images with very high spatial and spectral resolution than common machine learning methods. 

In our study, we exploit image clustering techniques on hyperspectral images to detect and evaluate the corrosion of the stones on cultural heritage assets. In more details, an automated mechanism is proposed for the detection of the ROIs in an unsupervised way. In other words, the evaluation of specific ROIs identifiability, using unsupervised clustering techniques, is being attempted.

\section{Related work}

A plethora of methods for assessing and detecting the deterioration of stone monuments are available to researchers \cite{grilli2019classification}. Those methods are distinguished into (a) destructive and (b) non-destructive techniques. The main drawback of the destructive approach is that a valuable  piece of monument structure is taken \cite{fitzner2002damage}. On the other hand, the non-destructive approaches utilize methods that extract features of the examined surface in order to detect cracks,  defects in the architectural surface and material degradation.  

In \cite{moropoulou2013non}, the authors proposed a method to identifying the exterior and interior surface flaws. This approach provides elastic features of the architectural structure materials in order to detect  the crack and inclusion in the building taking into consideration the affected layer inside the material.
In \cite{parida2018fuzzy}, the authors introduced a fuzzy clustering approach for extracting the local variance feature from an image. This method applied to define the transitional features implementing hybrid segmentation. 
In \cite{moropoulou2018multispectral}, the authors exploited infrared thermography to diagnose materials decay taking into account different historical periods.This approach is used as a tool in the diagnostic level, for the detection of invisible superficial cracks or/and disparities, as well as the revelation of moisture presence within structures.

In the case of cultural heritage, techniques such as clustering can be applied from archaeological artifacts to the entire archaeological site. 
In \cite{Pozo} the authors used multispectral images from different geomatic sensors, trying to define different construction materials and the main pathologies of Cultural Heritage elements by combining active and passive sensors recording data in different range. During this study, the unsupervised clustering method K-means is proposed. The results shows that an ideal sensors calibration can provide more accurate clustering. In \cite{APOLLONIO201889} the authors used 3D models and data mapping on 3D surfaces in the context of the restoration documentation of Neptune’s Fountain in Bologna. In \cite{oses2014image} the authors used machine learning classifiers, support vector machines and classification trees for the masonry classification. In \cite{messaoudi2018ontological} the authors developed a correlation pipeline for the integration of semantic, spatial and morphological dimension of a built heritage.

\begin{figure}
\includegraphics[width=0.9\linewidth]{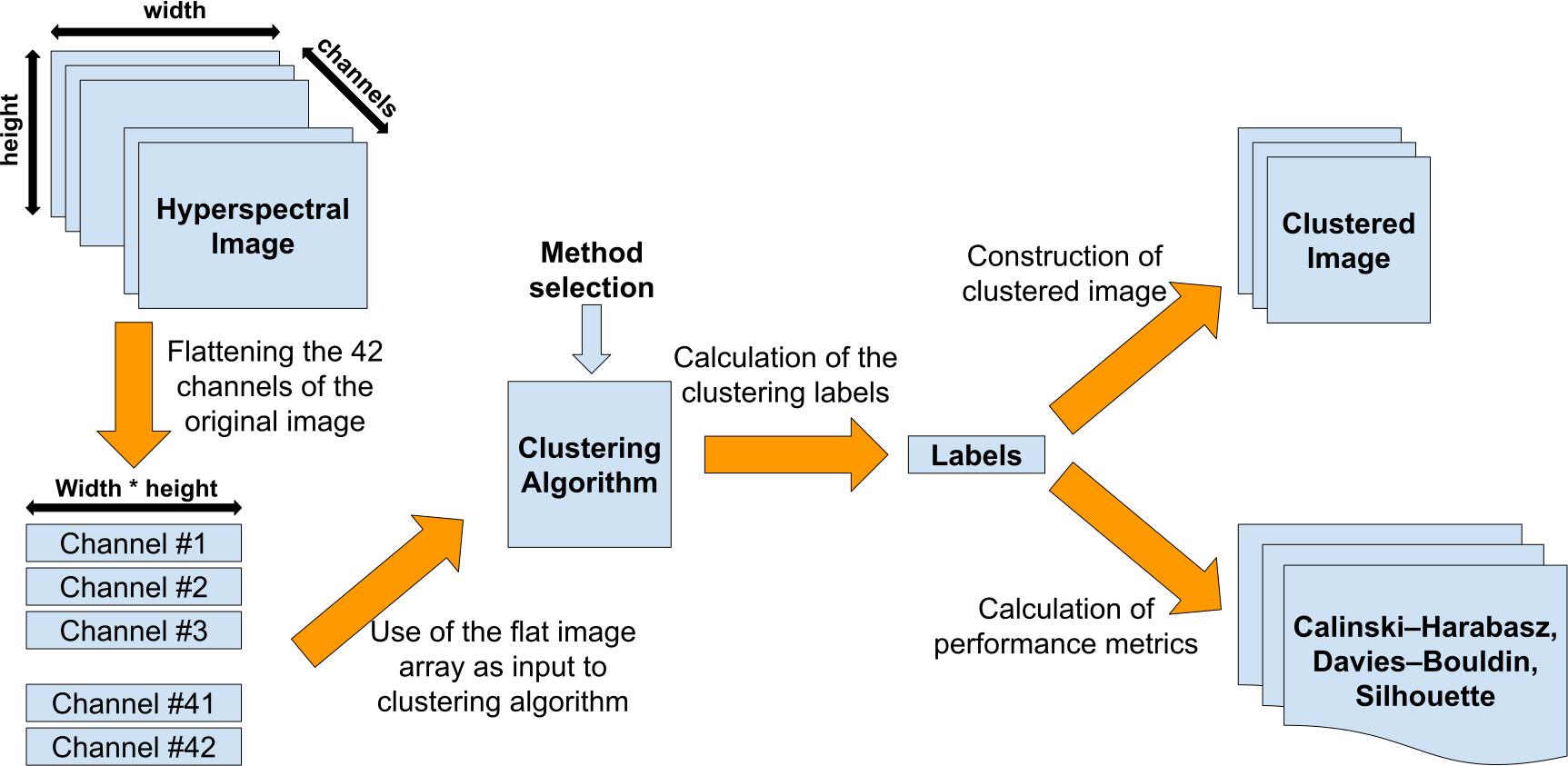} 
\caption{The proposed methodology workflow. Firstly, the original images are being converted to flat image arrays. At a second step, these flat images are used as input to clustering methods. Then, the clustering labels are produced. Finally, the clustered images are being constructed by the clustering labels and the performance metrics are being calculated for each clustering method.}
\label{fig:methodology}
\end{figure}

\section{Methodology Overview}

Cluster based machine learning approaches have been used for the detection of the wall corrosion, regarding the aforementioned historic monument. As described in \cite{HUANG2014293}, clustering is an unsupervised learning technique that is being applied to data in order to group them into clusters according to some common characteristics. Several well known clustering algorithms were applied to the images of this study like K-means, Meanshift, Spectral, Birch, DBSCAN and Optics \cite{Slawomir}.  

The data set, used for the purposes of this study, contains hyperspectral images with 42 channels, which represent the several frequencies of the electromagnetic spectrum. The first step of the proposed approach is the conversion of the initial images to flat image arrays, as shown in Figure \ref{fig:methodology}. At this stage, the image content is in the appropriate form defined by the clustering methods. The application of the clustering algorithms bring as result an array with labels that correspond to the produced clusters. Finally, these arrays of labels are used to build the so-called clustered images and to calculate the performance metrics of each clustering method.

\subsection{Our contribution}
The major outcome of this study is the development of an automated mechanism for the detection of several deterioration types on historical walls. To achieve this, a pipelined approach was followed for the decomposition of the initial images, the clustering fitting procedure, the construction of the clustered images and their comparison with the annotated ones. A pixel based processing was applied to the images, offering a more detailed analysis.  Another aspect of this study is the hyperspectral images selection as part of the dataset, since a correlation between the wall deterioration and the additional information from across the electromagnetic spectrum of the image was attempted.

\section{The experimental Setup}
Our proposed dataset consists of 14 final hyperspectral images of the  Fort of Saint Nicholas, with 42 channels for each image. These measurements carried out using the HyperView \cite{garea2016hyperview} multi sensor hyperspectral sensing platform by 3D-one. This HyperView system is a dual head system combining one Visual (VIS) snap-shot camera and one Near Infrared (NIR) snap-shot camera, which are connected on a EP-12 board. These cameras acquire only one band per pixel (instead of acquiring all spectral bands for every pixel) while they acquire all the bands in small windows, 4x4 for the VIS head and 5x5 for the NIR head. 

Then, the raw images turn into a low resolution hyperspectral image (1/4th or 1/5th of the initial resolution for VIS and the NIR camera respectively), an intermediate hyperspectral image  and the final pansharepened hyperspectral image.

\begin{figure}[h]

\begin{subfigure}{0.45\textwidth}
\includegraphics[width=0.9\linewidth, height=3cm]{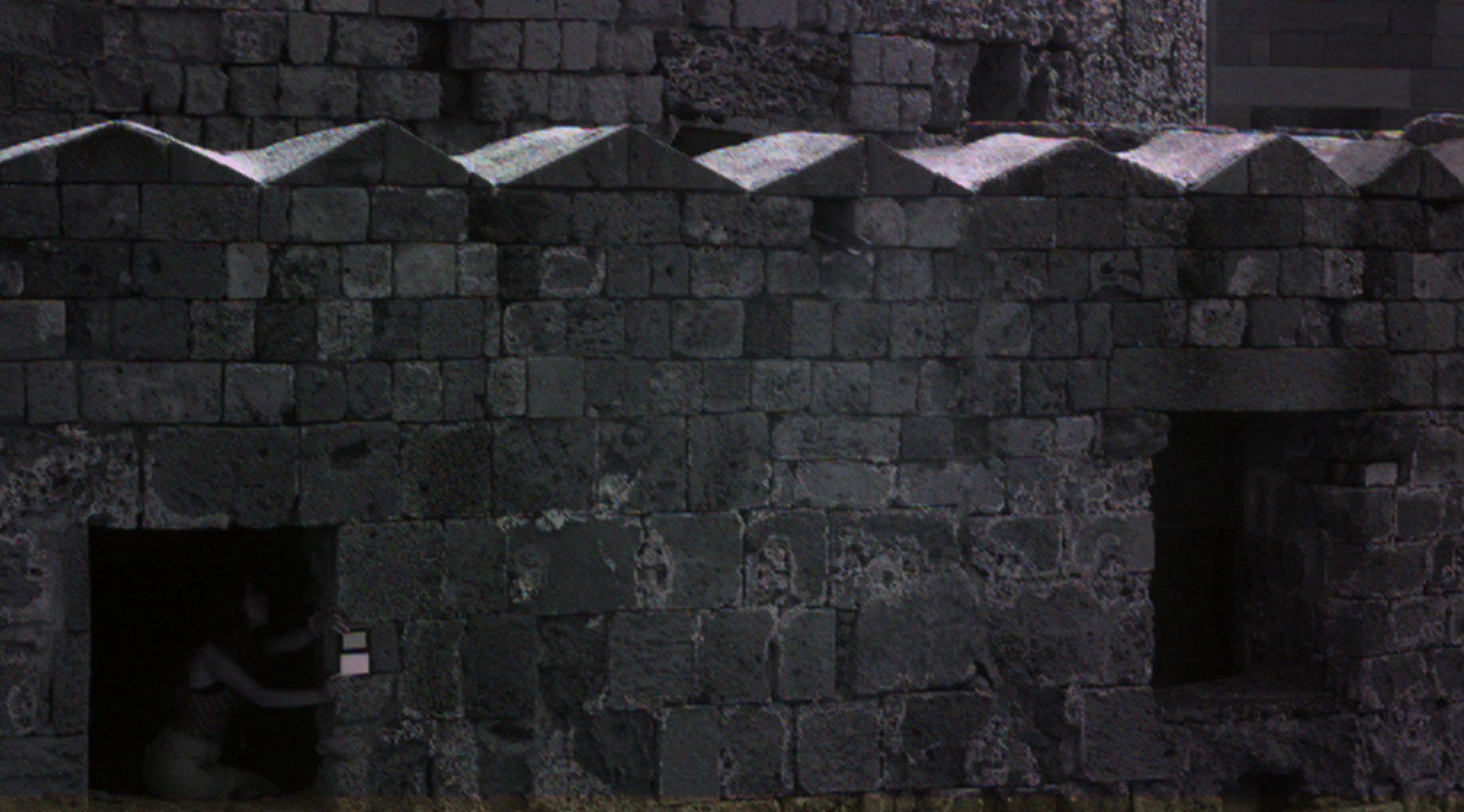} 
\caption{The original image}
\label{fig:nonannot}
\end{subfigure}
\begin{subfigure}{0.45\textwidth}
\includegraphics[width=0.9\linewidth, height=3cm]{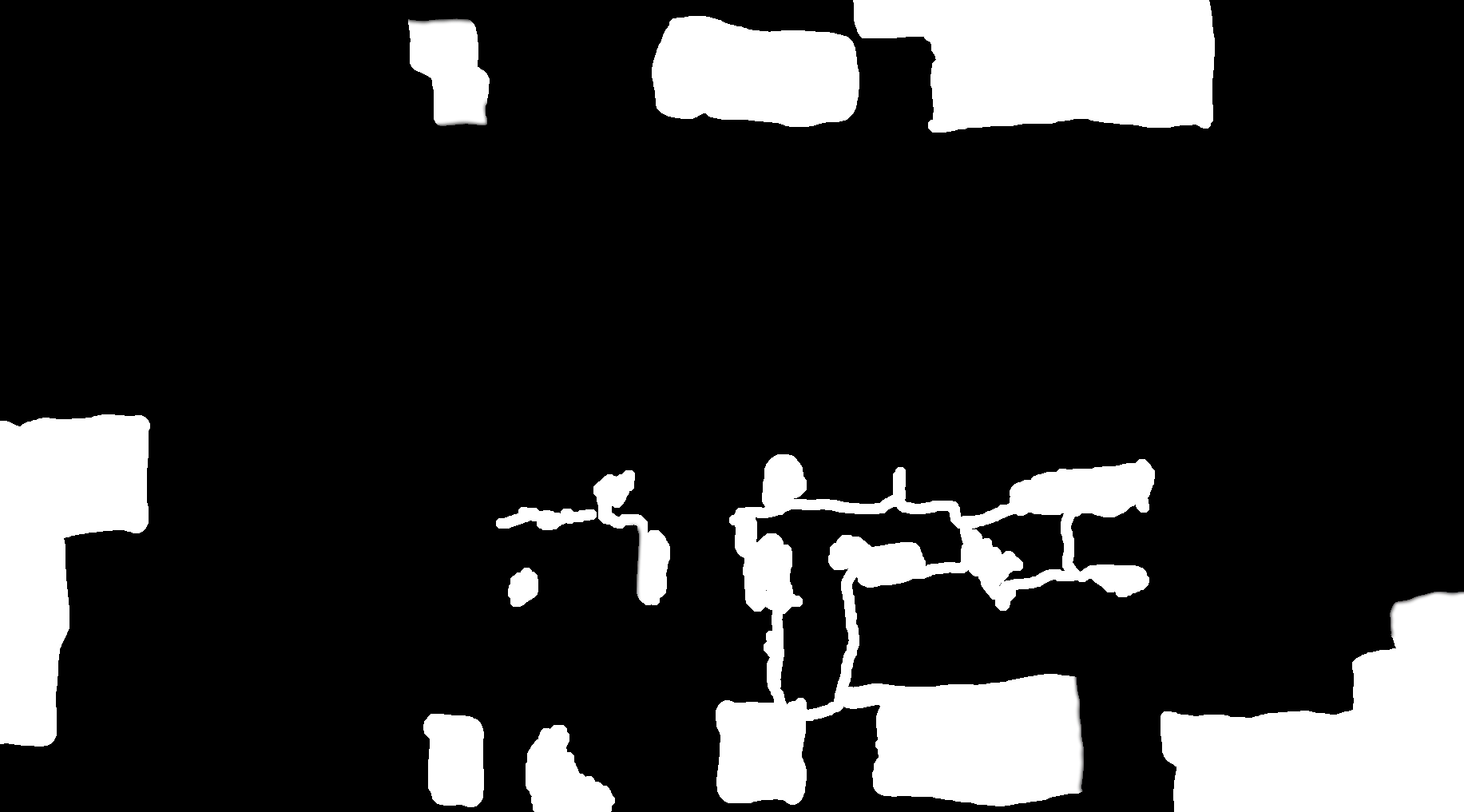}
\caption{The annotated image}
\label{fig:annotated}
\end{subfigure}
\caption{Less than 30\% of the wall is affected by the erosion. In the left image a small part of the wall looks brighter from the sunrays so is not considered as damaged.}
\label{fig:orivsanot}
\end{figure}

\subsection{The annotation process}
First step of the annotation process was the selection of the  hyperspectral image channels that correspond to the natural colours. The appropriate channel combination is the following triplet: 15th channel for Red, 6th channel for Green, and 3rd channel for Blue. Next step was the highlighting of the ROIs, using white colour. The rest of the image was coloured black in order for the damaged areas to be more clearly distinctive. 
In Figure \ref{fig:orivsanot}, the annotation process that implemented to one of the hyperspectral images, is presented. In the left photo (Figure \ref{fig:nonannot}) the image with the natural colours is shown, while in the right one (Figure \ref{fig:annotated}) the corresponding annotated image is depicted.

\subsection{Clustering areas depiction}
\begin{figure}[h]

\begin{subfigure}{0.45\textwidth}
\includegraphics[width=0.9\linewidth, height=3cm]{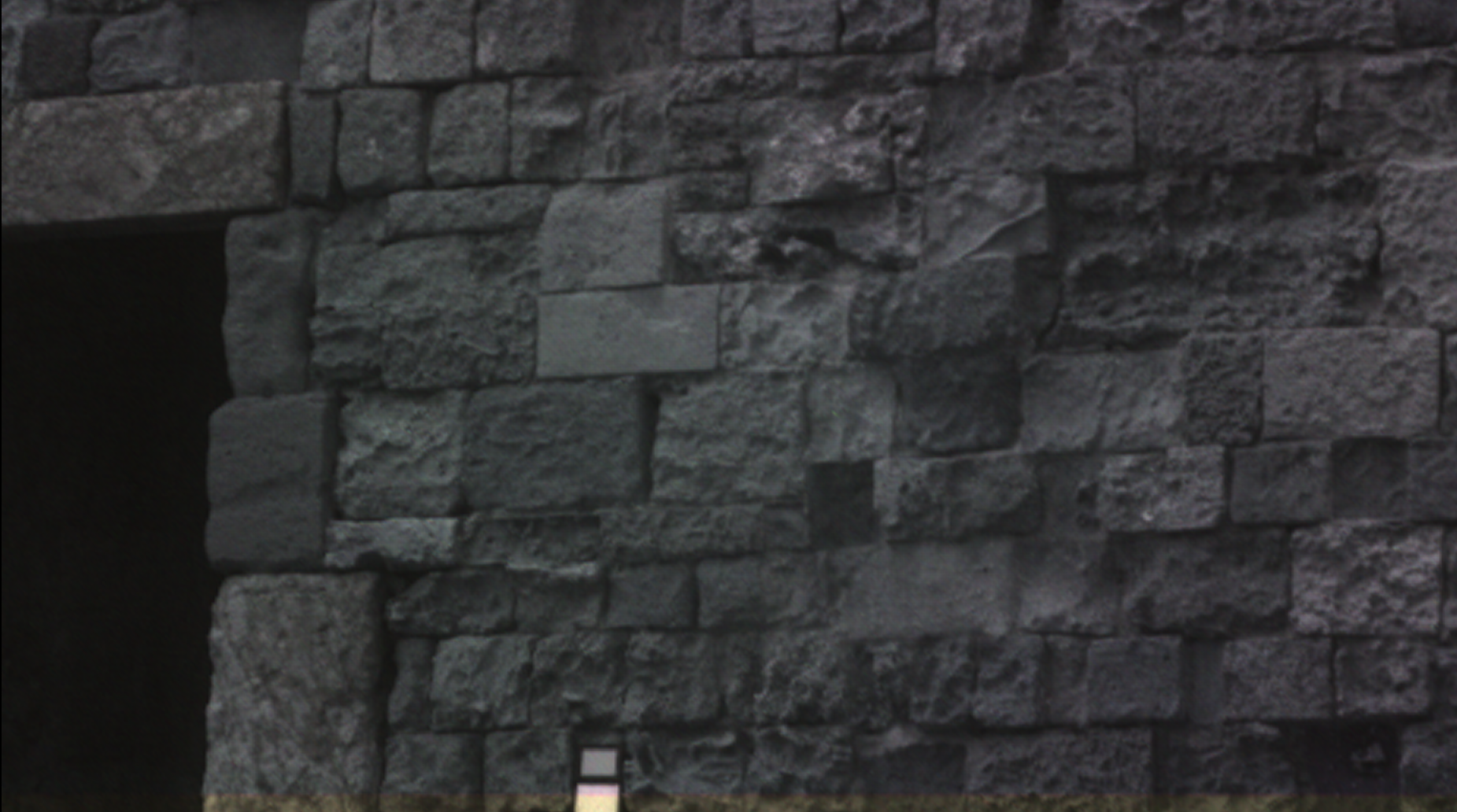} 
\caption{The original image}
\label{fig:original}
\end{subfigure}
\begin{subfigure}{0.45\textwidth}
\includegraphics[width=0.9\linewidth, height=3cm]{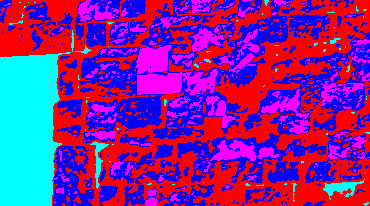}
\caption{The K-means clustering method}
\label{fig:K-means}
\end{subfigure}

\begin{subfigure}{0.45\textwidth}
\includegraphics[width=0.9\linewidth, height=3cm]{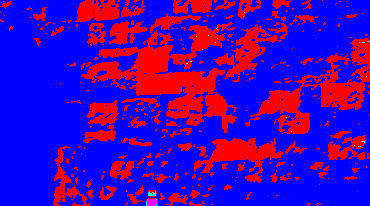} 
\caption{The Birch clustering method}
\label{fig:birch}
\end{subfigure}
\begin{subfigure}{0.45\textwidth}
\includegraphics[width=0.9\linewidth, height=3cm]{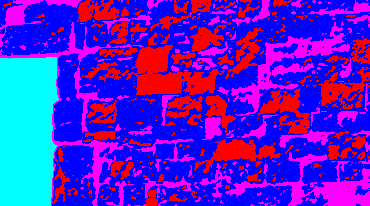}
\caption{The Spectral clustering method}
\label{fig:spectral}
\end{subfigure}

\caption{In Figure (a), the original hyperspectral image is presented, while in Figures (b), (c), (d) the images produced by methods K-means, Birch and Spectral are presented accordingly.}
\label{fig:clustering}
\end{figure}

The corrosion of the stones could be characterised by the different colour and the roughness of the relevant area surface on the corresponding images of the dataset. Thus, different clustering techniques could offer alternative views of the same data. 
In Figure \ref{fig:original}, the original image of the monument wall is depicted. Figure \ref{fig:K-means} presents the outcome of the K-means clustering algorithm, applied to that specific image, while Figures \ref{fig:birch}, \ref{fig:spectral} depict the Birch and the Spectral partitioning accordingly.

\section{Evaluation of Clustering methods}
\subsection{Clustering algorithms characterization}
Essential part of this study is the characterization of the several clustering techniques regarding their performance. Metrics such as Calinski–Harabasz \cite{Calinski}, Davies–Bouldin \cite{4766909} indexes and Silhouette value \cite{Papakostas} were calculated for the initial evaluation of the clustering methods. 

For the calculation of these metrics, some definitions and assumptions should be provided. Let K denotes the number of clusters $\{C_k\}, k=0, 1, 2, ..., K$. Let $X = \{x_1, x_2, ..., x_N\}$ be a vector containing N objects, where $x_{ij}$ denotes the jth element of $x_i$. The grouping of all objects $x_i, i=1,2,...,N$ in K clusters can be defined as follows:
\begin{equation} \label{eq:w}
w_{ki} = \begin{cases}
1, \hspace{0.5cm} iff x_i \in C_k \\
0, \hspace{0.5cm} otherwise.
\end{cases}
\end{equation} 
Eq. \ref{eq:w} ensures the uniqueness of the object to cluster association, which is a valid case for both hierarchical and partitioning cluster analysis

The Calinski–Harabasz index (CHI) is described by the Eq. \ref{eq:chi}:
\begin{equation}\label{eq:chi}
CHI(k) = \frac{T_B/(K-1)}{T_W/(K-1)}
\end{equation}

\begin{align*}
Where \hspace{0.2cm}T_B = \sum_{k=1}^{K}|\bar{C_K}|\lVert C_K-\bar{x}\rVert, \hspace{0.5cm}
T_W = \sum_{k=1}^{K}\sum_{i=1}^N w_{ki}\lVert x_i-\bar{C_K}\rVert ^2
\end{align*}

According to \cite{Papakostas}, the maximum CHI value is associated with the optimal partitioning of the given data. By using constant number of clusters for all clustering methods, the most fitting one gives the maximum CHI value. The Davies–Bouldin index (DBI) is an internal evaluation scheme, where the quality of the clustering is being examined according to information extracted directly from the given dataset. The DBI is defined by the Eq. \ref{eq:db}:
\begin{equation}\label{eq:db}
DB(k) = \frac{1}{K} \sum_{k=1}^K R_K
\end{equation}

\begin{align*}
where \hspace{0.2cm}R_K = max\Bigg(\frac{S_k+S_j}{d_{kj}}\Bigg), j=1,2, ..., K, j\neq K, \\
and \hspace{0.2cm} d_{kj} = \lVert \bar{x_k}-\bar{x_j}\rVert \label{eq:d}, \hspace{0.5cm} S_K = \frac{1}{\sum_{i=1}^{N}w_{ki}}\sum_{i=1}^{N}w_{ki}\lVert x_i-\bar{x_k}\rVert
\end{align*}

%\noindent where $R_K$ is defined as:

The minimum DBI value is related to the best partitioning solution. Thus, by using constant number of clusters for all clustering methods, the most fitting one gives the minimum DBI value. The silhouette value shows the similarity of an object regarding the cluster it belongs, compared to other clusters. The silhouette value is described by the Eq.\ref{eq:sil}
\begin{equation}\label{eq:sil}
s(x_i) = \frac{b(x_i)-a(x_i)}{max(b(x_i),a(x_i))}
\end{equation}
where $a(x_i)$ represents the average dissimilarity of the object with all the other data in the same cluster and $b(x_i)$ represents the lowest average dissimilarity of the object to any other cluster. Since, Silhouette value ranges from -1 to 1, a value close to 1 ensures that the object is well matched to its own cluster.

\subsection{Ground truth verification}
\begin{figure}[h]
\includegraphics[width=0.9\linewidth]{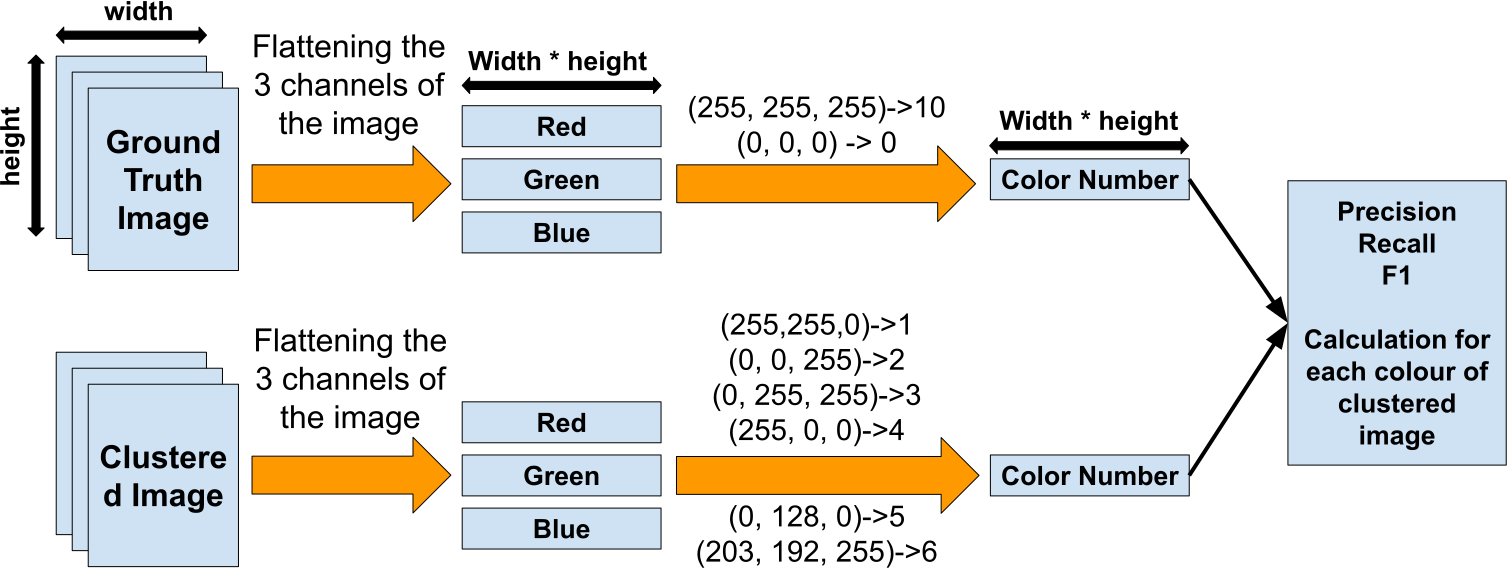} 
\caption{The clustered images evaluation. The annotated and the clustered images are being converted to flat RGB arrays. Each distinct RGB triplet gets associated to a unique identifier, representing the cluster label. The single dimensional arrays with the unique identifiers are combined for the calculation of accuracy, precision, recall and f1 scores}
\label{fig:groundTruth}
\end{figure}

To evaluate the outcome of the clustering methods, a set of annotated images was used that denote the ground truth regarding the corrosion area of the initial images. So, a comparison between the annotated and the clustered image was performed, using accuracy,  precision, recall and f1 scores \cite{rallis2018spatio}. The detailed procedure is depicted in Figure \ref{fig:groundTruth} and is distinguished into (a) conversion of annotated and clustered images to flat image arrays, (b) assignation of colour triplets (RGB) to specific identifiers that represent the clustering labels and the corresponding clustering colour, (c) accuracy, precision, recall and f1-scores calculation and (d) design of the corresponding graphs.

\begin{figure}[h]
\includegraphics[width=0.9\linewidth]{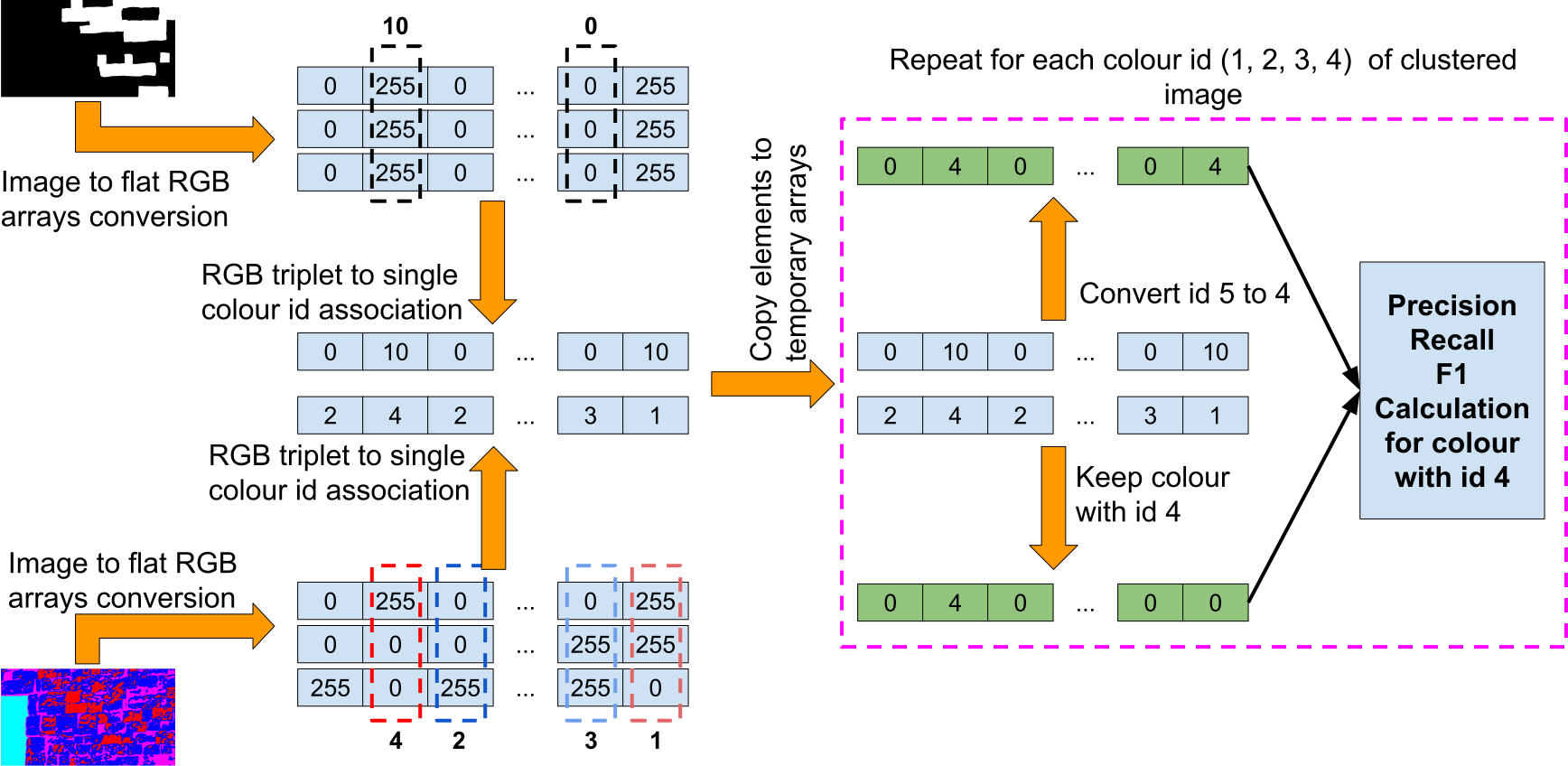} 
\caption{Results evaluation process. The annotated and the clustered images are being converted to flat RGB arrays. The annotated image RGB triplets get associated with identifiers 0 and 10, while the clustered image RGB triplets get associated with identifiers 1-6. These two single dimensional arrays are being adapted to the current clustering label identifier and the process is being repeated for each identifier.}
\label{fig:groundTruthComp}
\end{figure}

In Figure \ref{fig:groundTruthComp}, a simple example case is shown which offers a more descriptive view of the evaluation process. The initial images are being converted to RGB arrays. Each distinct RGB triplet is being assigned to a unique identifier that represents a specific cluster label (1-6) or one of the two distinct areas of the annotated images (0, 10). The two single-dimensional arrays are being adapted to the current clustering label which is under examination. Each identifier with number 10 of the annotated image single-dimensional array is being replaced by the identifier of the current clustering label. Each position of the clustered image single-dimensional array is being set to zero except from these which contain the same identifier of the current clustering label. The final single-dimensional arrays are used for the calculation of accuracy, precision, recall and f1 scores. This procedure is being repeated for each of the clustering labels (1-6). The most matching clustering label was extracted.

\section{Experimental results}
\begin{figure}[th]
\includegraphics[width=0.8\linewidth]{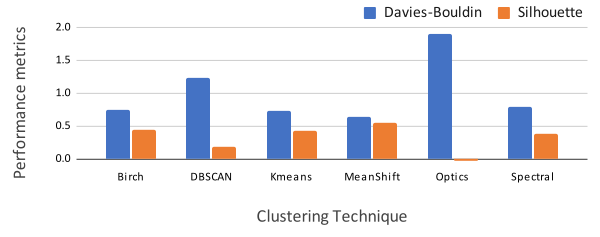} 
\caption{In this graph, the average cluster indexes (Calinski–Harabasz, Davies-Bouldin, Silhouette) for each clustering technique is presented. As shown, Meanshift method achieved the best performance, since its DBI is closest to 0 and Silhouette value is the closest to 1.}
\label{fig:db}
\end{figure}

As it was mentioned above, the initial evaluation of the clustring techniques was performed using the cluster indexes (Calinski–Harabasz, Davies-Bouldin, Silhouette), which are characterized as internal metrics. According to Davies-Bouldin metric, Meanshift clustering technique presents the best partitioning quality, since its value is the closest to 0. At the same time, Meanshift seems to achieve better similarity among the objects of a common cluster because the Silhouette value is closer to 1 than in any other case (Figure \ref{fig:db}).

\begin{figure}[th]
\includegraphics[width=0.8\linewidth]{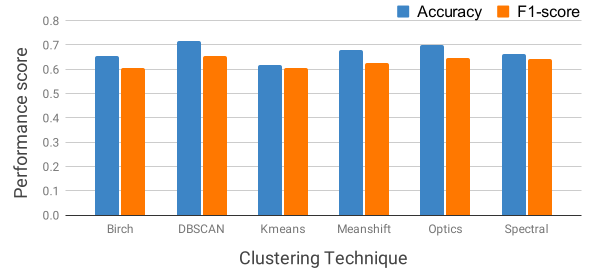} 
\caption{In this graph, the average accuracy and f1 performance scores are presented. As shown, DBSCAN achieved the best results.}
\label{fig:af}
\end{figure}

\begin{figure}[th]
\includegraphics[width=0.8\linewidth]{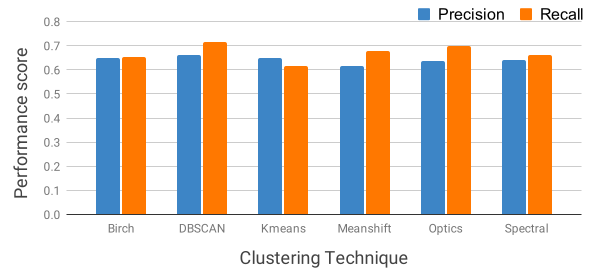} 
\caption{In this graph, the average precision and recall performance scores are presented. As shown, DBSCAN achieved the best results.}
\label{fig:pr}
\end{figure}

A secondary, more practical approach was used for the evaluation of the clustering methods by comparing the clustered images with the annotated ones and calculating the performance scores accuracy, precision, recall and f1 (Figures \ref{fig:af}, \ref{fig:pr}). From this evaluation, it arises that DBSCAN was the most fitting technique, since it achieved the best scores. Consequently, despite the better internal performance metrics of Meanshift, DBSCAN proved to be the technique that identified more sufficiently the ROIs of the given images.  


\section{Conclusion}
In our approach, we investigated whether the spectral signatures suffice to distinct various ROIs  using trivial unsupervised machine learning techniques. Therefore, we investigate various clustering approaches to identify the feasibility of such methods.
According to the results, unsupervised techniques can provide an early, still appropriate mechanism regarding the identification of certain regions of an image (monitoring, defect recognition).

\section*{Acknowledgement}
This paper is supported by the European Union Funded  project Hyperion "Development of a Decision Support System for Improved Resilience \& Sustainable Reconstruction of historic areas to cope with Climate Change \& Extreme Events based on Novel Sensors and Modelling Tools" under the Horizon 2020 program H2020-EU.3.5.6., grant agreement No 821054. 

\bibliographystyle{unsrt}
\bibliography{euuchm}

\begin{thebibliography}{10}

\bibitem{rallis2017extraction}
Ioannis Rallis, Ioannis Georgoulas, Nikolaos Doulamis, Athanasios Voulodimos,
  and Panagiotis Terzopoulos.
\newblock Extraction of key postures from 3d human motion data for choreography
  summarization.
\newblock In {\em 2017 9th international conference on virtual worlds and games
  for serious applications (VS-Games)}, pages 94--101. IEEE, 2017.

\bibitem{moropoulou2013non}
Antonia Moropoulou, Kyriakos~C Labropoulos, Ekaterini~T Delegou, Maria
  Karoglou, and Asterios Bakolas.
\newblock Non-destructive techniques as a tool for the protection of built
  cultural heritage.
\newblock {\em Construction and Building Materials}, 48:1222--1239, 2013.

\bibitem{marrie2008unesco}
Henrietta Marrie.
\newblock The unesco convention for the safeguarding of the intangible cultural
  heritage and the protection and maintenance of the intangible cultural
  heritage of indigenous peoples.
\newblock In {\em Intangible heritage}, pages 183--206. Routledge, 2008.

\bibitem{adamopoulos2017multi}
E~Adamopoulos, E~Tsilimantou, V~Keramidas, M~Apostolopoulou, M~Karoglou,
  S~Tapinaki, C~Ioannidis, A~Georgopoulos, and A~Moropoulou.
\newblock Multi-sensor documentation of metric and qualitative information of
  historic stone structures.
\newblock {\em ISPRS Annals of Photogrammetry, Remote Sensing \& Spatial
  Information Sciences}, 4, 2017.

\bibitem{forsyth2012computer}
David~A Forsyth and Jean Ponce.
\newblock {\em Computer vision: a modern approach}.
\newblock Pearson, 2012.

\bibitem{maltezos2018building}
Evangelos Maltezos, Anastasios Doulamis, Nikolaos Doulamis, and Charalabos
  Ioannidis.
\newblock Building extraction from lidar data applying deep convolutional
  neural networks.
\newblock {\em IEEE Geoscience and Remote Sensing Letters}, 16(1):155--159,
  2018.

\bibitem{makantasis2015deep}
Konstantinos Makantasis, Konstantinos Karantzalos, Anastasios Doulamis, and
  Nikolaos Doulamis.
\newblock Deep supervised learning for hyperspectral data classification
  through convolutional neural networks.
\newblock In {\em 2015 IEEE International Geoscience and Remote Sensing
  Symposium (IGARSS)}, pages 4959--4962. IEEE, 2015.

\bibitem{grilli2019classification}
Eleonora Grilli and Fabio Remondino.
\newblock Classification of 3d digital heritage.
\newblock {\em Remote Sensing}, 11(7):847, 2019.

\bibitem{fitzner2002damage}
Bernd Fitzner.
\newblock Damage diagnosis on stone monuments-in situ investigation and
  laboratory studies.
\newblock In {\em Proceedings of the International Symposium of the
  Conservation of the Bangudae Petroglyph}, volume~7, pages 29--71, 2002.

\bibitem{parida2018fuzzy}
Priyadarsan Parida and Nilamani Bhoi.
\newblock Fuzzy clustering based transition region extraction for image
  segmentation.
\newblock {\em Engineering Science and Technology, an International Journal},
  21(4):547--563, 2018.

\bibitem{moropoulou2018multispectral}
Antonia Moropoulou, Nicolas~P Avdelidis, Maria Karoglou, Ekaterini~T Delegou,
  Emmanouil Alexakis, and Vasileios Keramidas.
\newblock Multispectral applications of infrared thermography in the diagnosis
  and protection of built cultural heritage.
\newblock {\em Applied Sciences}, 8(2):284, 2018.

\bibitem{Pozo}
Susana Del~Pozo, Pablo Rodríguez-Gonzálvez, Luis Sánchez-Aparicio,
  A.~Muñoz-Nieto, David Hernandez, Beatriz Felipe, and Diego
  González-Aguilera.
\newblock Multispectral imaging in cultural heritage conservation.
\newblock {\em ISPRS - International Archives of the Photogrammetry, Remote
  Sensing and Spatial Information Sciences}, XLII-2/W5:155--162, 08 2017.

\bibitem{APOLLONIO201889}
Fabrizio~Ivan Apollonio, Vilma Basilissi, Marco Callieri, Matteo Dellepiane,
  Marco Gaiani, Federico Ponchio, Francesca Rizzo, Angelo~Raffaele Rubino,
  Roberto Scopigno, and Giorgio Sobra’.
\newblock A 3d-centered information system for the documentation of a complex
  restoration intervention.
\newblock {\em Journal of Cultural Heritage}, 29:89--99, 2018.

\bibitem{oses2014image}
Noelia Oses, Fadi Dornaika, and Abdelmalik Moujahid.
\newblock Image-based delineation and classification of built heritage masonry.
\newblock {\em Remote Sensing}, 6(3):1863--1889, 2014.

\bibitem{messaoudi2018ontological}
Tommy Messaoudi, Philippe V{\'e}ron, Gilles Halin, and Livio De~Luca.
\newblock An ontological model for the reality-based 3d annotation of heritage
  building conservation state.
\newblock {\em Journal of Cultural Heritage}, 29:100--112, 2018.

\bibitem{HUANG2014293}
Xiaohui Huang, Yunming Ye, Huifeng Guo, Yi~Cai, Haijun Zhang, and Yan Li.
\newblock Dskmeans: A new kmeans-type approach to discriminative subspace
  clustering.
\newblock {\em Knowledge-Based Systems}, 70:293--300, 2014.

\bibitem{Slawomir}
Mieczyslaw~Klopotek Slawomir~Wierzchon.
\newblock {\em Modern Algorithms of Cluster Analysis}.
\newblock Springer International Publishing, 2018.

\bibitem{garea2016hyperview}
Alberto~S Garea, {\'A}lvaro Ord{\'o}{\~n}ez, Dora~B Heras, and Francisco
  Arg{\"u}ello.
\newblock Hyperview: an open source desktop application for hyperspectral
  remote-sensing data processing.
\newblock {\em International Journal of Remote Sensing}, 37(23):5533--5550,
  2016.

\bibitem{Calinski}
Tadeusz Caliński and Harabasz JA.
\newblock A dendrite method for cluster analysis.
\newblock {\em Communications in Statistics - Theory and Methods}, 3:1--27, 01
  1974.

\bibitem{4766909}
David~L. Davies and Donald~W. Bouldin.
\newblock A cluster separation measure.
\newblock {\em IEEE Transactions on Pattern Analysis and Machine Intelligence},
  PAMI-1(2):224--227, 1979.

\bibitem{Papakostas}
Papakostas~George A.
\newblock Stacked autoencoders for outlier detection in over-the-horizon radar
  signals.
\newblock {\em Computational Intelligence and Neuroscience}, 322, 2017.

\bibitem{rallis2018spatio}
Ioannis Rallis, Nikolaos Doulamis, Anastasios Doulamis, Athanasios Voulodimos,
  and Vassilios Vescoukis.
\newblock Spatio-temporal summarization of dance choreographies.
\newblock {\em Computers \& Graphics}, 73:88--101, 2018.

\end{thebibliography}

\end{document}